\documentclass[lettersize,article,twoside]{IEEEtran}
\usepackage{amsmath,amsfonts}
\usepackage{array}
\usepackage[caption=false,font=normalsize,labelfont=sf,textfont=sf]{subfig}
\usepackage{textcomp}
\usepackage{url}
\usepackage{verbatim}
\usepackage{graphicx}
\usepackage{cite}

\usepackage{xcolor}
\usepackage[bookmarks=true]{hyperref}
\usepackage{threeparttable} 
\usepackage{makecell}
\usepackage{hhline}
\usepackage{amsmath}  %
\usepackage[Symbol]{upgreek}%
\usepackage{bm}%
\usepackage{graphicx}
\usepackage{caption}
\usepackage{subfig}
\usepackage{epsfig}
\usepackage{mathrsfs}
\usepackage{amsfonts,amssymb}
\usepackage{algpseudocode}
\usepackage{cite}
\usepackage{supertabular}
\usepackage{longtable}
\usepackage{hhline}
\usepackage{multirow}
\usepackage{algorithm}
\usepackage[normalem]{ulem}
\usepackage{dblfloatfix}    
\usepackage[bottom]{footmisc} 
\usepackage{tikz}

\usepackage{changes}
\usepackage[T1]{fontenc}
\usepackage{aecompl}

\def\BibTeX{{\rm B\kern-.05em{\sc i\kern-.025em b}\kern-.08em
    T\kern-.1667em\lower.7ex\hbox{E}\kern-.125emX}}

\begin{document}
\bibliographystyle{ieeetr} 
\title{ The Developments and Challenges towards Dexterous and Embodied Robotic Manipulation: A Survey }

\author{Gaofeng Li,~\IEEEmembership{Member,~IEEE,}
        Ruize Wang,~\IEEEmembership{Student Member,~IEEE,}
        Peisen Xu,~\IEEEmembership{Student Member,~IEEE,}\\
        Qi Ye,~\IEEEmembership{Member,~IEEE,}
        and Jiming Chen,~\IEEEmembership{Fellow,~IEEE}
\thanks{This work is supported by the Independent Project of State Key Laboratory of Industrial Control Technology under Grant ICT2025A06, Open Competition Mechanism on Humanoid Robot of Longyou County, Zhejiang Province, Major Scientific and Technological Research Project under Grant 2025246, Special Fund for Innovation and Development of Hangzhou West Science and Technology Innovation Corridor under Grant K-20241950, and the National Natural Science Foundation of China (NSFC) under Grant 62203426.}

\thanks{All authors are with the College of Control Science and Engineering, Zhejiang University, Hangzhou, 310027, China, and with the Key Laboratory of Collaborative Sensing and Autonomous Unmanned Systems of Zhejiang Province. (e-mails: {\{gaofeng.li, ruize.wang, peisen.xu, qi.ye, cjm\}}@zju.edu.cn). G. Li is the corresponding author.}}

\markboth{IEEE ROBOTICS \& AUTOMATION MAGAZINE}%
{Li \MakeLowercase{\textit{et al.}}: The Developments and Challenges towards Dexterous and Embodied Robotic Manipulation: A Survey}


\maketitle

\begin{abstract}
Achieving human-like dexterous robotic manipulation remains a central goal and a pivotal challenge in robotics. The development of Artificial Intelligence (AI) has allowed rapid progress in robotic manipulation. This survey summarizes the evolution of robotic manipulation from mechanical programming to embodied intelligence, alongside the transition from simple grippers to multi-fingered dexterous hands, outlining key characteristics and main challenges. Focusing on the current stage of embodied dexterous manipulation, we highlight recent advances in two critical areas: dexterous manipulation data collection (via simulation, human demonstrations, and teleoperation) and skill-learning frameworks (imitation and reinforcement learning). Then, based on the overview of the existing data collection paradigm and learning framework, three key challenges restricting the development of dexterous robotic manipulation are summarized and discussed.

\end{abstract}


\section{Introduction}
\IEEEPARstart{M}{aking} robots have the humanoid dexterous manipulation ability is the most essential and central goal in the development of robotics\cite{science.aat8414}. However, in contrast to the great achievements of Artificial Intelligence (AI) in imitating humans' cognition and learning abilities, the robotic dexterous manipulation dedicated to imitating humans' interactions with the physical environments has progressed at a relatively slower pace. Robotic manipulation refers to the behavior of a robot that controls its own actuators to interact with the environment, thereby affecting the physical world \cite{2020-MIT-robotic-manipulation}. Since the birth, robots are expected to have humanoid manipulation abilities. However, limited by the perception and cognition ability, robots are only largely applied in structured industrial scenes for basic and repeated pick-and-place tasks.

\begin{figure}[]
    \centerline{\includegraphics[width=1\linewidth]{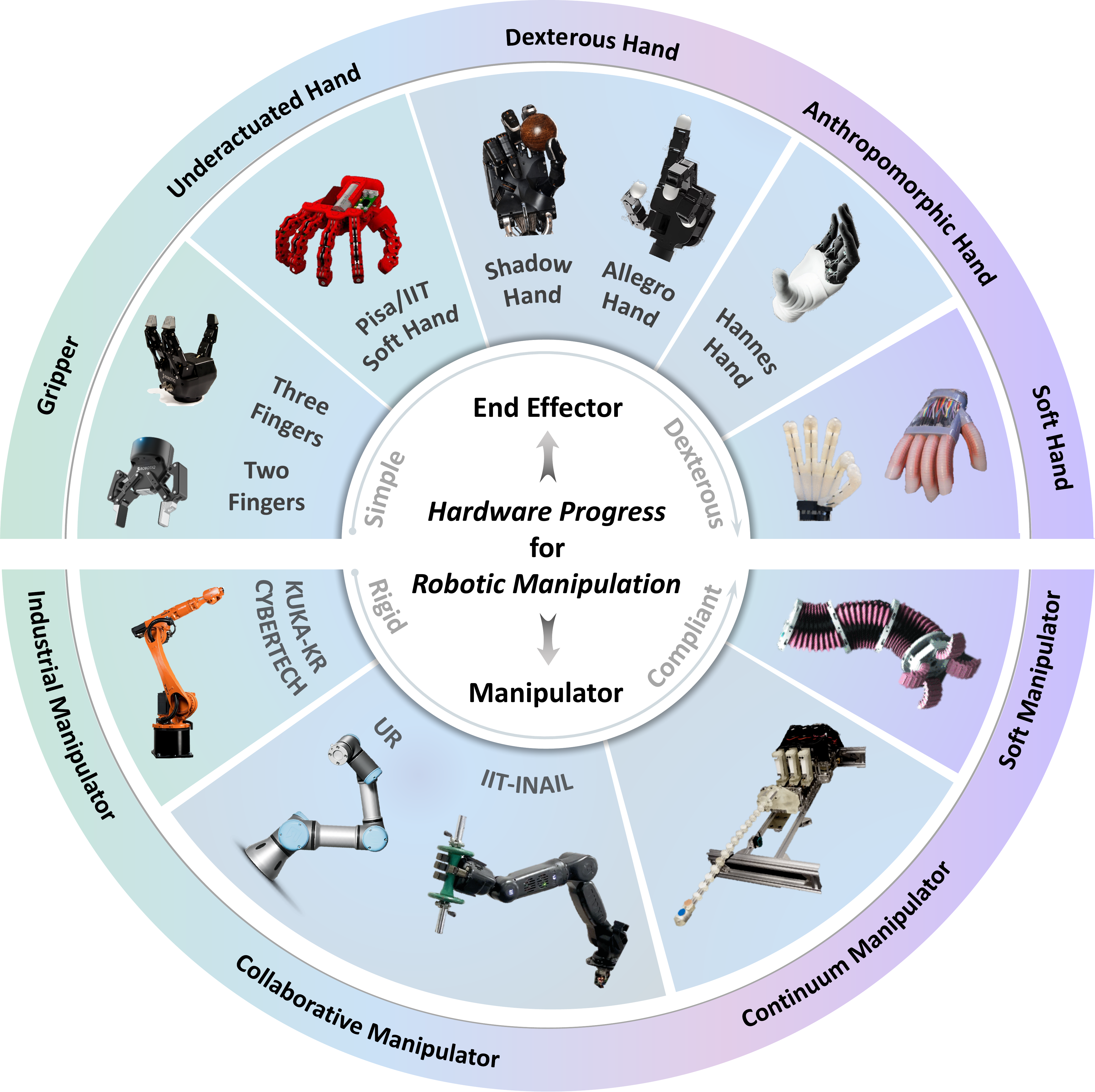}}
    \caption{The hardware progress of end-effector and manipulator for robotic manipulation.}
    \label{hand_development}
\end{figure}

In recent years, robotics technology has made great progress. To lay a solid foundation for robotic dexterous manipulation, it is necessary not only for the robot to become more compliant and stable, but also for the end-effector to be more dexterous. As shown in Fig. \ref{hand_development}, the robot has gradually developed from the traditional industrial manipulator to the collaborative manipulator\cite{2021AIM}, which can interact with humans and the uncertain environment. Meanwhile, there are also some specially designed robots such as the continuum manipulators \cite{2016RAL-Continuum-Manipulators}, which can continuously deform to the desired shape and the soft manipulator\cite{2020RAM-Soft-Arm} made of special soft materials. 
The robot's end-effector has also evolved from the simplest parallel gripper to multi-fingered grippered\cite{2014franchi-technical}, soft underactuated hands\cite{2018TRO-Dexterous-Manipulation}, rigid dexterous hands\cite{2019ICRA-Shadow-Dexterous-Hand}, and highly anthropomorphic hands with fully-actuated design\cite{scirobotics.abb0467}. At the same time, with the development of soft robotics technology, some soft grippers and soft hands have gradually developed to mature\cite{2023TRO-Soft-Robotic-Gripper}\cite{2016SR}. These advances have provided a solid hardware foundation for robotic dexterous manipulation, greatly improving the robot's manipulation capabilities. 

However, even today, robotic manipulation is still an unresolved and open problem. The performance of robotic manipulation is still far from humans' expectation, especially when it comes to hand's dexterous manipulation\cite{2016ISAM-Baxter-research-robot}\cite{2017IndustrialRobot-domestic}. {\color{black}On one hand, the dexterous hands have high degrees of freedom (DoFs), making it difficult to achieve precise control through models. On the other hand, the learning-based methods need a large amount of high-quality datasets for training, especially the human demonstration data which can significantly accelerate the training process.} Therefore, robotic manipulation is still facing numerous challenges and is a hot topic at the cutting edge of the robotics field. 

In order to find the optimal solution to realize dexterous robotic manipulation, Liu {\it et al.}\cite{2021Robotics-review} summarized recent progress of deep reinforcement learning (DRL\textit{}) algorithms in tackling the application problems for robotic manipulation control. Furthermore, Mohammed {\it et al.}\cite{2022Sensors-review} specifically divided DRL-based robotic manipulation tasks in cluttered environments into three categories for review. Fang {\it et al.}\cite{2019Springer-survey} reviewed the art of imitation learning (IL) for robotic manipulation from three aspects including demonstration, representation, and learning algorithms. However, these reviews only focus on the isolated learning algorithms and lack a more comprehensive summary on the developments and challenges of robotic manipulation.

Yu {\it et al.}\cite{2022Frontiers-review-dexterous} provided a comprehensive review of the methods for dexterous manipulation with multi-fingered robotic hands from the model-based method to the latest research based on reinforcement learning (RL). 
Han {\it et al.}\cite{2023Sensors-survey} summarized recent advances in learning-based methods for robotic manipulation tasks. They examined the problems that arise when applying these algorithms to robotic manipulation tasks and the various solutions that have been proposed to deal with these problems. Although they have summarized the methods for robotic manipulation comprehensively, they focused too much on the algorithms to implement robotic manipulation itself and neglected the manipulation data collection for training, which is the important cornerstone to achieve satisfied effectiveness of learning-based methods. Liu {\it et al.}\cite{2019AASreview} divided the latest methods for learning robotic manipulation skills into three categories based on the different collection of datasets to review. But the method of data collection is still not involved. Newbury {\it et al.}\cite{2023TRO-survey} provided a systematic review of deep learning methods in robotic object grasping including learning methods, datasets, benchmarking, and so on. However, they mainly summarized the use of simple two- or three-fingered grippers, with few considering anthropomorphic hands, and focused more on dataset design rather than data collection method. Thus, a more systematic overview is needed not only to provide a comprehensive exploration of the recent advance in robotic dexterous manipulation learning, but also to analyze the developments and challenges of robotic manipulation to provide a clearer path for future development.

Our main contributions are summarized as follows. 
\begin{itemize}
    \item[$\bullet$] In chronological order, we summarize the development process of robotic manipulation and divide it into three stages: Mechanical Programming Stage, Closed-Loop Control Stage, and Embodied Intelligent Manipulation Stage. Meanwhile, we give the characteristics of each stage and analyze the challenges in achieving dexterous robotic manipulation.
    \item[$\bullet$] Focusing on the current stage of embodied dexterous manipulation, we systematically list recent advances from two aspects: data collection methods and manipulation skill learning framework. 
    \item[$\bullet$] Based on the analysis on existing methods, we conclude three key challenges of dexterous robotic manipulation, which we consider crucial for future research to realize human-like dexterous manipulation.
\end{itemize}

The rest of this paper is organized as follows. Section \ref{Section-II} summarizes the development of robotic manipulation. Section \ref{Section-III} analyzes the challenges in current stage. Sections \ref{Section-IV} and \ref{Section-V} review current advances in data collection method and learning framework of dexterous manipulation, respectively. The typical scenarios and their advantages/disadvantages and open issues are also discussed. Section \ref{Section-VI} summaries the challenges in the current development process of robotic manipulation. Section \ref{Section-VII} concludes this survey.


\section{The Historical Stages of Robotic Manipulation}
\label{Section-II}

\begin{figure}[]
    \centerline{\includegraphics[width=1\linewidth]{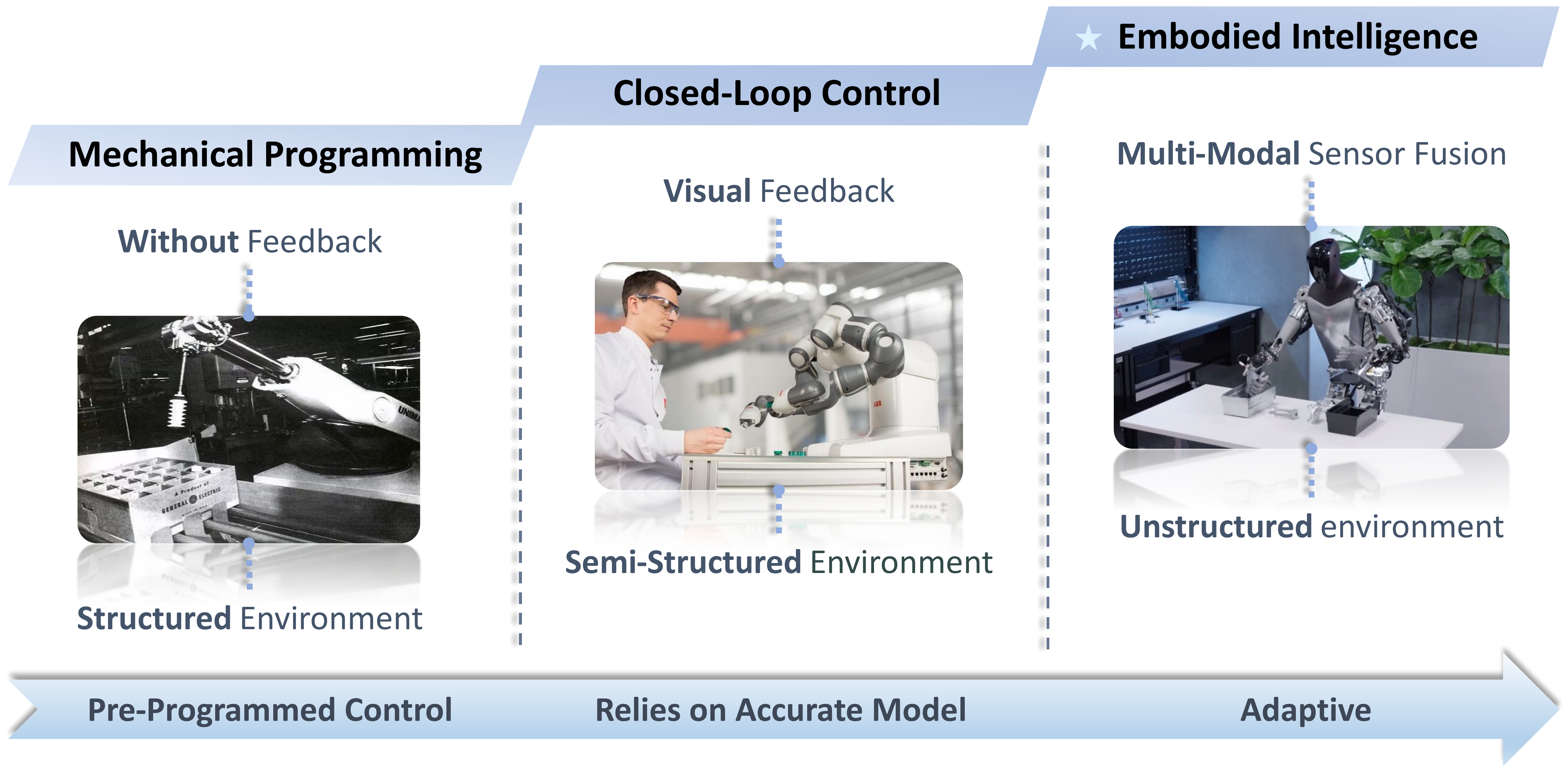}}
    \caption{The development process of robotic manipulation. There are three stages, including mechanical programming, closed-loop control, and embodied intelligence.}
    \label{development}
\end{figure}

In recent years, a consensus has been reached that the disembodied intelligence without a physical body will eventually reach a ceiling, and the embodied intelligence with physical entities (e.g., robots) to interact with the world is the only way to achieve general AI\cite{2022TETCI-Survey}. {\color{black}Meanwhile, the robotic manipulation has entered the stage of embodied intelligence. This evolution stems not only from advances in AI and robotic hardware, but also from growing performance expectations for robotic manipulation itself. To better understand its characteristics and pathway to embodied intelligence, this section examines the developmental history of robotic manipulation. The development can be divided into three stages as shown in Fig. \ref{development}.}

The first stage is the {\bf{Mechanical Programming Stage}}. Industrial robots, e.g., the Unimate and PUMA560 with a parallel gripper can realize pick-place manipulation on industrial production lines under pre-programmed controller \cite{2021ICICT-Minimum}. However, robots at this stage do not have external sensing ability and lack the ability to adapt to external changes (such as changes in the position and shape of the workpiece). 

The second stage is the {\bf{Closed-Loop Control Stage}} based on visual servo. The visual feedback is introduced into the control loop through the Eye-in-Hand or Eye-to-Hand camera to achieve closed-loop control based on feature tracking. This enables robotic manipulation to have a certain degree of adaptability to changes in tasks and environments. Typical achievements such as the ABB YuMi dual-arm collaborative robot \cite{2020application-le}, which can complete component assembly tasks in a semi-structured environment. However, the control of the robot at this stage still relies on accurate modeling of the external environment or the operated workpiece. For example, in a model-based grasping task, force closure detection is a necessary condition for determining whether a stable grasp can be formed. However, the inaccuracy of the model will greatly reduce the manipulation performance, while the unstructured environment is always filled with a large number of objects and disturbances that are difficult to be modeled in advance. 

With the development of AI, robotic manipulation has gradually entered the {\bf{Embodied Intelligent Manipulation Stage}} that emphasizes the end-to-end "perception-decision-execution" closed loop \cite{2021SR-toward}. {\color{black}This paradigm represents a fundamental shift from previous stages. On the perception front, robots no longer rely solely on visual feedback but instead leverage multi-modal sensor fusion, integrating vision, force, and tactile sensing. Force and tactile feedback provide critical information about physical interactions, enabling manipulation in unstructured environments where visual feedback alone are insufficient. On the decision front, robotic manipulation is gradually breaking free from the constraints of precise environmental modeling. Robot agents can now autonomously decide their next actions and trajectories based on real-time multi-modal perceptual feedback, even in previously unknown environments. On the execution front, robots are increasingly equipped with multi-fingered dexterous hands as end-effectors, which are capable of acquiring complex manipulation skills and achieving far more flexible and adaptive interactions with objects.} Embodied intelligent manipulation enables robots to enhance their understanding of the surrounding environment based on multi-modal sensor information such as vision, force, and tactile, and is expected to enable robots to adapt to dynamic unstructured environments. It is currently the most promising technical route to give robots the human-like dexterous manipulation abilities.


\section{The Challenges in Dexterous Manipulation: from Gripper to Muti-Fingered Hand}
\label{Section-III}

\begin{figure*}[]
    \centerline{\includegraphics[width=1\linewidth]{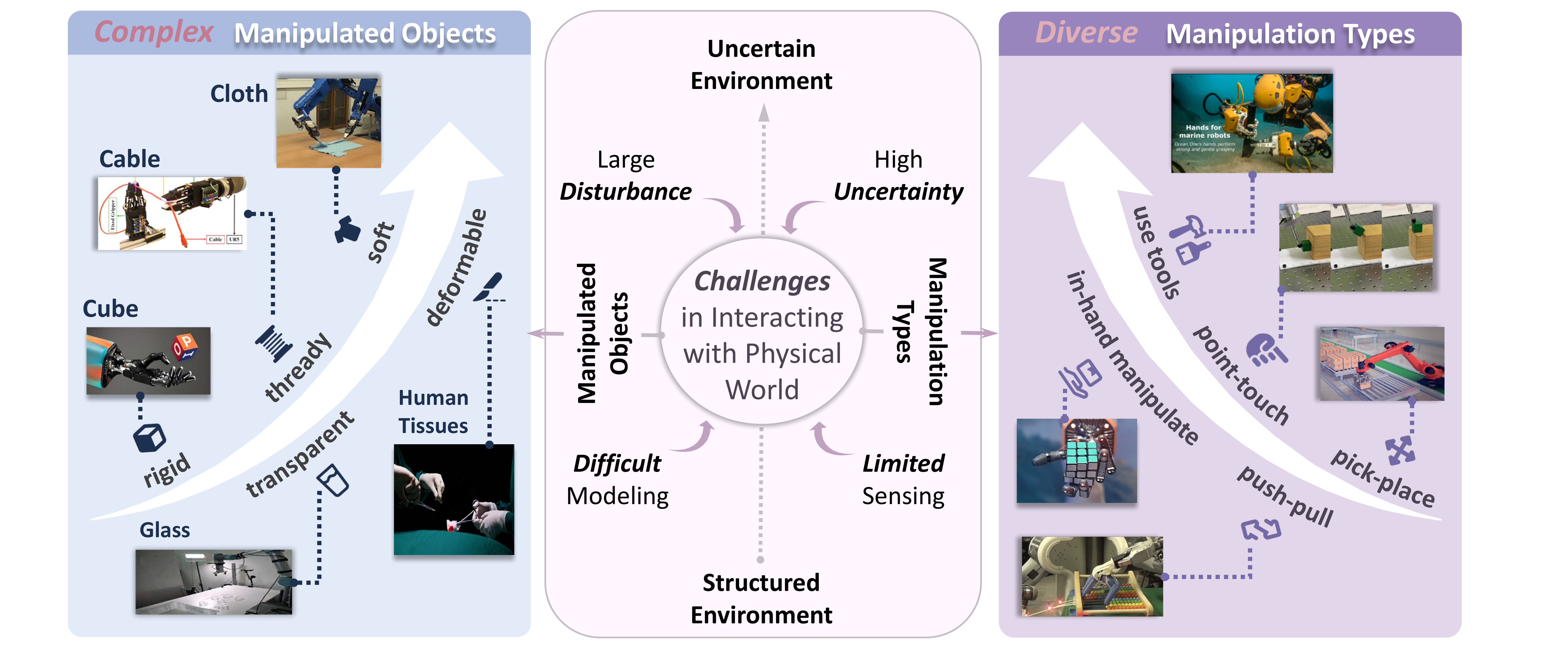}}
    \caption{The challenges in interacting with physical world. The robot has to face more and more complex manipulated objects and diverse manipulation types.}
    \label{chanllenge}
\end{figure*}

In the embodied intelligent manipulation stage, the complexity of interacting with the physical world is the main source of the severe challenges faced by robotic manipulation. As robots move beyond structured factory environments, the objects manipulated by robots have also changed from traditional rigid workpieces to complex objects that are difficult to be sensed or modeled. As shown in Fig. \ref{chanllenge}, it is challenging for robots to manipulate linear objects (e.g., cables)\cite{2023TRO-Global-Model-Learning}, transparent objects (e.g., glasses)\cite{2022RAL-A4T}, soft objects (e.g., cloth)\cite{2018IJRR-Robotic-manipulation} \cite{2019IJARS-review}, and deformable objects (e.g., human tissues in medical operations)\cite{2020IROS-soft}. 
When robots interact with the physical world, except for the more and more complex manipulated objects, the types of robotic manipulation have also become more diverse. In addition to the basic pick-and-place manipulation, a lot of robotic manipulation types are required for interaction, such as point-touch sliding manipulation that relies on single-point contact to achieve complex object movement\cite{2021ICRA-Contact-model}, push-pull manipulation (e.g. turning an abacus and opening/closing switches)\cite{2016IROS-learning}, in-hand manipulation that only relies on one hand to achieve object rotation\cite{2022RAL-complex}, and dexterous manipulation in various scenarios (e.g., opening/closing valves, grasping objects, and using tools)\cite{2022RAM-challenges}. Complex objects and diverse manipulations make robotic manipulation face challenges in difficult modeling, high uncertainty, large disturbances, and limited sense capabilities, which greatly affect the further expansion and application of robot technology.

How to solve these problems is the frontier and hot topic in current research. However, current research on robotic manipulation mainly focuses on two-fingered grippers. For example, the robotic large language model (LLM) RT-X series proposed by Google \cite{2024ICRA-open} and VoxPoser \cite{2023arxiv-voxposer} which combines LLM and vision-language model (VLM) for robotic manipulation. Although a simple two-fingered gripper can finish many manipulation tasks, this type of end-effector has two disadvantages which seriously restrict its development in robotic manipulation. First, the two-fingered gripper can only provide fewer contact points during the interaction and cannot provide stable and reliable gripping under certain tasks. Second, the two-fingered gripper itself has fewer degrees of freedom and cannot complete complex manipulation tasks, such as in-hand rotating and using tools. 

Highly anthropomorphic multi-fingered dexterous hands can make up the deficiency of two-fingered grippers. {\color{black}On the one hand, it has more DoFs, contact points, and a larger operating space, which enhance adaptability and stability in uncertain environments and when handling objects difficult to model. On the other hand, dexterous hands support a wider range of skills instead of only opening and closing as grippers. The dexterous hands can perform human-like manipulations such as twisting and screwing, and autonomously select actions based on tactile feedback \cite{2024IROS_DexSkills}. Although grippers can also be equipped with tactile sensors, they can only detect basic grasp status rather than enabling active environmental interaction. By contrast, dexterous hands can use rich tactile sensing to compensate for limited visual perception (e.g., blocked cameras) and interact adaptively with the environment. Thus, only dexterous hands are capable of addressing all challenges in robotic manipulation.

However, learning manipulation skills for multi-fingered hands remains significantly more difficult than for simple two-fingered grippers.} Compared with the simple two-fingered gripper, the learning of manipulation skills for multi-fingered dexterous hands is particularly difficult. First, the multi-fingered dexterous hand has a higher degree of freedom (DoF) and a more variable interaction space, which greatly increases the difficulty of searching in high-dimensional space for algorithms such as RL. Second, in the process of performing manipulation tasks, the multi-fingered dexterous hand has multiple contact points with objects, leading to more complex and diverse mechanical properties. This further makes the learning of skills for multi-fingered dexterous hands more difficult.

Endowing the multi-fingered dexterous hand with human-like dexterous manipulation capabilities is important to promote the development of robotics. As one of the important organs for human movement execution, most of the human daily work must use their dexterous hands, such as dressing, carrying plates, and picking up food. Lepora \cite{2024SR-future} also pointed out that "the future lies in a pair of tactile hands". Therefore, endowing robotic hands with human-like dexterous manipulation capabilities can enhance the application potential of robots, especially humanoid robots, to perform complex tasks. This can promote further improvement of productivity by robotics and AI, profoundly change human production and lifestyle, and reshape the pattern of global industrial development. 
At the same time, there are also some studies showing that the evolution of human arms and hands has greatly promoted the evolution of the human brain, which is one of the key factors for humans to evolve into advanced animals. Therefore, studying the skill learning of robotic multi-fingered dexterous hands can enhance the understanding of the decision-making mechanism of the human brain and greatly promote the development of the robot's "cerebellum" and "brain". Thus, research on the learning of dexterous manipulation skills for multi-fingered dexterous hands is of great significance and can promote the early arrival of strong AI.



\section{The Data Collection Paradigm for Dexterous Manipulation}
\label{Section-IV}
\begin{table*}[htbp]
\setlength{\tabcolsep}{4pt} 
\renewcommand{\arraystretch}{1.6} 
\centering
\caption{Data Collection Paradigms of Dexterous Manipulation}
\begin{threeparttable}
{
    \label{table1}
    \begin{tabular}{m{80pt}<{\centering}|m{60pt}<{\centering}|m{20pt}<{\centering}|m{60pt}<{\centering}|m{250pt}<{\centering}}
    \hline
    \multicolumn{1}{c}{Data Collection Method} &\multicolumn{1}{c}{Name} &\multicolumn{1}{c}{Ref.} &\multicolumn{1}{c}{End-Effector} &\multicolumn{1}{c}{Features}\\
    
    \hline
    \multirow{4}{*}{} & GraspM$^3$ & \cite{2024icra-tpgp} & Multi-Fingered Hand & Provide millions of grasping trajectories covering 8,152 objects with Shadow Hand.\\
    \cline{4-5}
    \multirow{4}{*}{\makecell{Simulation \\ Platform}} & RoboGen & \cite{2024icml} & Gripper & A generative robotic agent with automatically generating diversified tasks, scenes, and training supervisions.\\
    \cline{4-5}
    \multirow{4}{*}{} & DexGraspNet & \cite{2023ICRA-dexgraspnet}& Multi-Fingered Hand & Contain 1.32 million grasping data of 5,355 objects in 133 categories for Shadow Hand.\\
    \cline{4-5}
    \multirow{4}{*}{} & GRPtopia & \cite{2024arxiv-grutopia} & -- & Provide 89 functional scenarios and 100,000 high-quality interactive data for robot training.\\
    
    \cline{1-5}
    \multirow{3}{*}{\makecell{Human \\ Demonstration}} & Videodex & \makecell{\cite{2023CRL-videodex} \\ \cite{2023CVPR-affordances}}& Multi-Fingered Hand &  Leverage visual, action, and physical priors from human video datasets to guide robot behavior.\\
    \cline{4-5}
    \multirow{3}{*}{} & VTDexManip & \makecell{\cite{liu2025vtdexmanip} \\ \cite{2024icra-Visual-Tactile}} & Multi-Fingered Hand & Collect a vision-tactile dataset by humans manipulating 10 daily tasks and 182 objects.\\
    \cline{4-5}
    \multirow{3}{*}{}&  {\color{black}ActionSense} & {\color{black}\cite{delpreto2022actionsense}} & {\color{black}Multi-Fingered Hand} & {\color{black} An expanding dataset focused on daily tasks in a kitchen environment featuring wearable sensors for body and finger tracking, forearm muscle activity, tactile information, and eye tracking with first-person video.}\\
    
    \cline{1-5}
    \multirow{5}{*}{} & Open-TeleVision & \cite{pmlr-2025-cheng25b} & Multi-Fingered Hand & Mirror the operator's arm and hand movements on the robot through VR devices without force feedback.\\
    \cline{4-5}
    \multirow{5}{*}{\makecell{Teleoperation \\ Demonstration}} & GR00T & \cite{2025arxiv-gr00t} & Multi-Fingered Hand & Support multiple devices to capture human hand motion for teleoperation demonstration data collection without force feedback.\\
    \cline{4-5}
    \multirow{5}{*}{} & DexCap  & \cite{2024arXivDexCap} & Multi-Fingered Hand & Capture human 6-DoF poses of the wrist and the finger motions through mocap glove and multi-camera without force feedback.\\
    \cline{4-5}
    \multirow{5}{*}{} & Mobile ALOHA & \cite{2024arXivmobileALOHA} & Gripper & Collect bimanual mobile manipulation data through a whole-body teleoperation system.\\
    \cline{4-5}
    \multirow{5}{*}{} & $\pi0$ & \cite{2024arXivpi0} & Gripper & Collect data in 68 tasks composed of complex behaviors on both single- and dual-arm robots with low control frequency.\\
    \cline{1-4}

    \hline
    
    \end{tabular}
} 
\end{threeparttable} 
\vspace{-5pt}
\label{data collection method}
\end{table*}

From the mechanical programming stage, the closed-loop control stage to the embodied intelligent stage, robotic manipulation has achieved remarkable results with the support of AI. Massive high-quality dataset is an important cornerstone for the current framework of AI, which is based on the deep neural network. When AI expands from disembodied intelligence limited in cyberspace to embodied intelligence that emphasizes interaction with the real physical world, how to obtain massive amounts of interactive data becomes the key to embodied intelligence. As shown in Table. \ref{data collection method}, there are three paradigms for collecting massive interactive data, namely: data generation based on simulation platform, data collection from human demonstration, and data collection from teleoperation demonstration.

\subsection{Data Generation Based on Simulation Platform}

Faced with complex, diverse, and highly uncertain application scenarios, traditional learning methods that rely on "robot-environment" interaction to obtain intelligence often require millions of iterations to learn useful skills, which is generally inefficient. Therefore, the most common approach is to generate a large amount of data through simulation platforms to improve efficiency. Data generation based on simulation platforms has some unique advantages. First, generating data on the simulation platform can avoid the high cost and potential danger of repeated experiments in the real world. 
Second, the simulation platform can efficiently, cost-effectively and repeatably generate a large amount of pedigree data with comparative significance. This can enhance the diversity of datasets and improve the quality of generated data. Finally, the simulation platform can accelerate the speed of data collection that may take longer in the real physical world by changing the flow rate of time in simulation environment. 

At present, there have developed some simulation datasets based on physics engines such as Genesis\cite{Genesis}, Isaac Sim, PyBullet\cite{PyBullet2021}, and MoJoCo\cite{todorov2012mujoco}. For example, the GraspM$^3$\cite{2024icra-tpgp} simulation dataset that provides millions of grasping trajectories covering 8,152 objects and a generative robot agent RoboGen\cite{2024icml} that can generate data infinitely. Wang {\it et al.}\cite{2023ICRA-dexgraspnet} generated a large-scale simulation dataset DexGraspNet for Shadow Hand, which contains 1.32 million grasping data of 5,355 objects in 133 categories. Shanghai Artificial Intelligence Laboratory\cite{2024arxiv-grutopia} released a city-level embodied intelligent simulation platform GRPtopia, which can provide 89 functional scenarios and 100,000 high-quality interactive data for robot training. 

{\color{black}Meanwhile, the tactile feedback serving as the crucial information for dexterous manipulation has also attracted increasing attention in data generation and collection. Wang {\it et al.} \cite{2022ral_TACT0} designed a simulator, named TACTO, for vision-based tactile sensors. This simulator allows to render realistic high-resolution touch readings in dexterous manipulation at hundreds of frames per second, and can be easily configured to simulate different vision-based tactile sensors. Zhong {\it et al.} \cite{2025TRO_TactGen} proposed a TactGen for cross-modal visuo-tactile data generation, which can achieve zero-shot sim-to-real transfer without any real-world data for domain adaptation. Hong {\it et al.} \cite{2024CVPR_MultiPLY} collected a large-scale multisensory interaction dataset, named Multisensory Universe, comprising 500k data by deploying an LLM-powered embodied agent to engage with the 3D environment. Despite progress in generating tactile data, great potential remains for creating a dexterous manipulation simulation dataset with tactile sensing.}

These works greatly promote the improvement of robotic manipulation capabilities, especially grasping capabilities. However, the shortcomings of this paradigm are also very obvious. First, there are various deviations between the simulation environment and the real environment, such as inaccurate friction modeling and air resistance modeling. This will inevitably introduce the Simulation-to-Real (Sim2Real) gap, making the models trained by simulation data be difficult to be directly migrate to real robots. Secondly, for the simulation of some complex objects, such as deformable objects and flexible objects, the performance of the current physics engine is still unsatisfactory. Moreover, the calculation of these objects often requires a lot of computing resources and computing time. Therefore, the paradigm of generating data based on the simulation platform is still difficult to replace the data collected from the real world.

\subsection{Data Collection from Human Demonstration}

The ultimate goal of robotic manipulation is to endow robots with human-like dexterous manipulation capabilities. Therefore, learning manipulation skills directly from human demonstration actions has become an attractive solution. Although the scale of data collection based on human motion demonstrations cannot compare with the data generated by simulation, it significantly reduces the difficulty compared to repeated trials with real robots, and can expand the scope of data collection without requiring robot hardware. At the same time, compared with the data generated by simulation, the human demonstration data is real interaction data produced in the physical environment, which can greatly reduce the Sim2Real gap. 

For instance, Bahl {\it et al.}\cite{2023CRL-videodex}\cite{2023CVPR-affordances} proposed to extract human motions and key interaction elements from billions of video data that are already widely available in Internet to achieve manipulation data collection based on human demonstration. Liu {\it et al.}\cite{liu2025vtdexmanip}\cite{2024icra-Visual-Tactile} also developed a vision-tactile fusion human motion capture system, constructing the VTDexManip dataset. 
{\color{black} DelPreto {\it et al.} \cite{delpreto2022actionsense} built a wearable data collection framework for recording synchronized multimodal data through body trackers, first-person video with attention estimates, muscle activity sensors, and custom tactile gloves. They used this system to collect an expanding dataset focused on daily tasks in a kitchen environment. Zhao {\it et al.} \cite{2024RAL-Tactile-Based-Grasping} used a wearable tool-like parallel hand exoskeleton with visual-based tactile sensors to collect data from human demonstration by grasping 15 different daily objects.} These works enhance robots’ manipulation capabilities in real physical environments.

\begin{figure}[t!]
    \centerline{\includegraphics[width=0.75\linewidth]{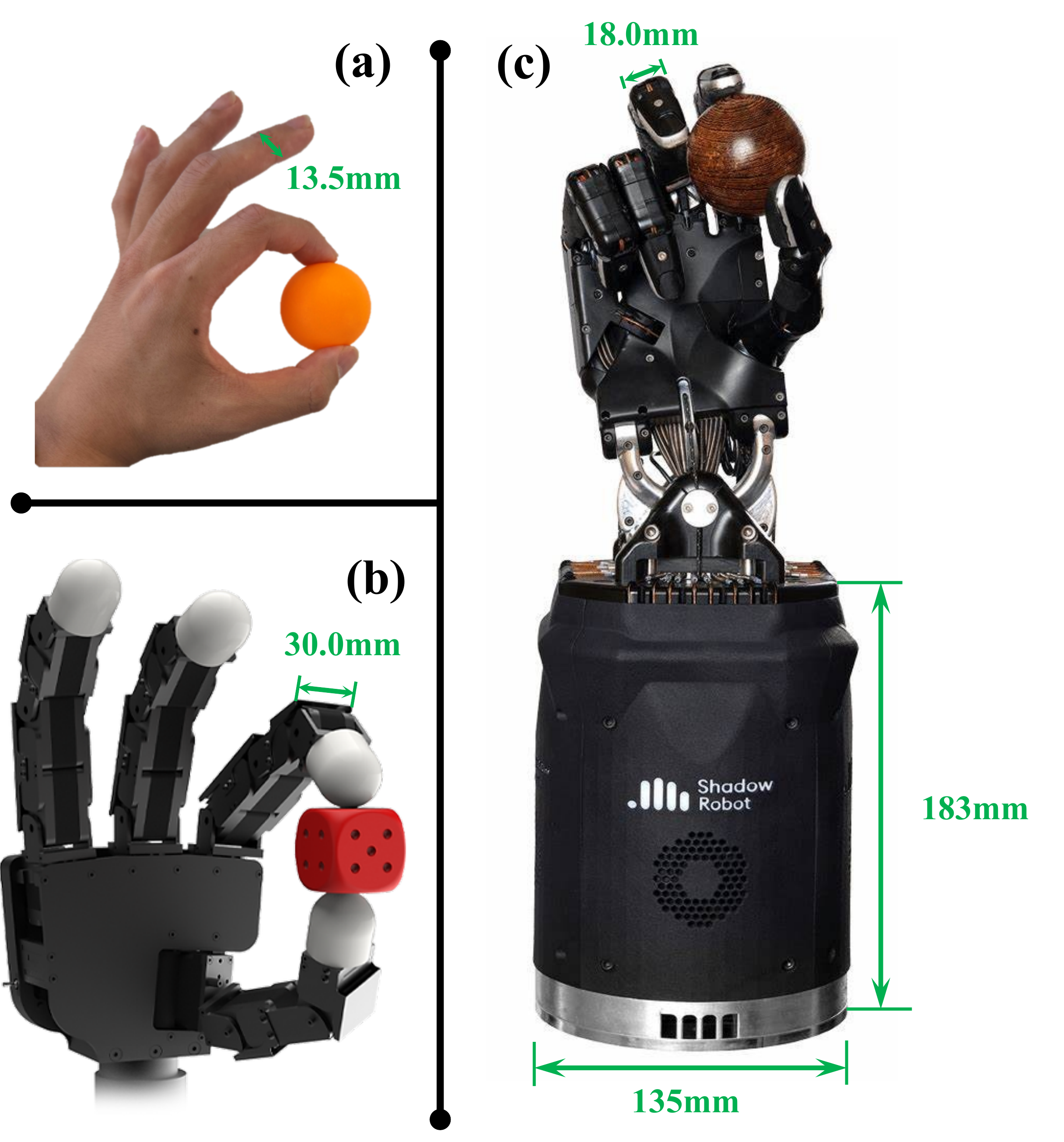}}
    \caption{Illustration of the human-to-robot gap. The comparison of (a) human hand, (b) Allegro Hand and (c) Shadow Hand. It is obvious that the Allegro Hand's fingers are mach larger than human and the Shadow Hand has a huge driven box, which bring negative effects to the Arm-Hand system.}
    \label{hand_compare}
\end{figure}

However, the current mechatronic systems still pale in comparison to the human musculoskeletal system in terms of drive-transmission efficiency and energy density. Consequently, significant morphological differences persist between robotic systems and the human body, particularly between human hands and robotic hands. Achieving high DoFs comparable to human hands within limited spaces remains highly challenging. The current robotic dexterous hand and the human hand are still very different in configuration. For example, the Allegro Hand has only four fingers and each finger is much larger than the human finger. Although the fingers of the Shadow Hand are as slender as human hands, there is a huge driven box at the back, which severely limits the reachable space of the hand. The comparison of those multi-fingered dexterous hands and human hand is shown in Fig. \ref{hand_compare}. It is pretty obvious that there is a huge gap in finger size and structure between human hand and multi-fingered dexterous hand. 
Due to the existence of these differences between human and robot, it is difficult to directly reproduce human demonstration data to robots, resulting in the Human-to-Robot gap.

\subsection{Data Collection from Teleoperation Demonstration}

To solve the Sim2Real gap and Human-to-Robot gap, the telerobotic system becomes a more efficient solution. Telerobotic systems can integrate human intelligence into robotic manipulation through Human-Robot shared control\cite{2023-TH}, while strictly adhering to the robot’s inherent kinematic/dynamic constraints. This can effectively mitigate the Sim2Real and Human-to-Robot gaps inherent in other data collection paradigms. Several teams have developed data acquisition systems based on telerobotic systems in recent years. However, existing systems still face the following problems. First, most of the current telerobotic systems are "weakly coupled" systems based only on visual feedback, lacking force and tactile feedback to humans. For example, the Open-TeleVision system\cite{pmlr-2025-cheng25b} co-developed by MIT and UCSD, NVIDIA's GR00T\cite{2025arxiv-gr00t} system utilizing Vision Pro, and the DexCap system\cite{2024arXivDexCap}. However, in some contact-rich tasks, human force and tactile expertise struggles to effectively integrate into the collected data\cite{2021TMech-angle}\cite{2020RAL-novel}. Secondly, most of the current telerobotic systems are based on two-fingered grippers, and lack high DoFs datasets from multi-fingered dexterous hands, such as Mobile ALOHA\cite{2024arXivmobileALOHA} and $\pi0$\cite{2024arXivpi0}. However, how to achieve accurate motion capture and force feedback for human hands with over 20 DoFs in confined spaces remains a challenging problem\cite{2024ICRA-joint}. Unlike traditional telerobotic systems that only involve end-effector force interactions, multi-fingered dexterous teleoperation introduces multi-point contacts and complex dynamic interactions between robotic fingers and objects. This complexity creates new challenges in system stability and transparency analysis, urgently requiring novel theoretical frameworks and analytical methods. Finally, existing teleoperation systems exhibit latencies exceeding tens of milliseconds, and the system agility is far from enough. As one of the earliest research directions in robotics, traditional telerobotic systems often focus on dealing with latency problems caused by long-distance communication\cite{2023RAS-telerobotic}. However, data collection for embodied intelligent manipulation puts forward new requirements for the agility of telerobotic systems. How to ensure the agility of the telerobotic system remains an understudied critical issue\cite{2023TH-haptic}\cite{2021RAL-workspace}.


\section{The Learning Framework for Dexterous Manipulation Skills}
\label{Section-V}
The frameworks for robotic manipulation skill learning primarily fall into two categories: imitation learning (IL) and reinforcement learning (RL). This section examines the current state of research in both paradigms and the challenges they face. As shown in Table. \ref{IL and RL}, the features of different IL and RL are listed.

\begin{table*}[htbp]
\setlength{\tabcolsep}{7.5pt} 
\renewcommand{\arraystretch}{1.6} 
\centering
\caption{Learning Framework of Dexterous Manipulation Skills}
\begin{threeparttable}
{
    \label{table2}
    \begin{tabular}{m{45pt}<{\centering}|m{55pt}<{\centering}|m{20pt}<{\centering}|m{300pt}<{\centering}}
    \hline
    \multicolumn{1}{c}{Category} &\multicolumn{1}{c}{Method / Name} &\multicolumn{1}{c}{Ref.} &\multicolumn{1}{c}{Features}\\
    
    \hline
    \multirow{15}{*}{\makecell{Imitation \\ Learning}} & DMP & \cite{2013-neuralcomp-learningAttractor} & Encode demonstration trajectories from IL into stable and reproducible attractor dynamics.\\
    \cline{4-4}
    \multirow{15}{*}{} & SEDS & \cite{2011TRO-LearningStable} & Model motion as a nonlinear autonomous dynamical system.\\
    \cline{4-4}
    \multirow{15}{*}{} & ProMP & \cite{2013NIPS-ProbabilisticMovementPrimitives} & Model the variability of motion through probabilistic models.\\
    \cline{4-4}
    \multirow{15}{*}{} & TPGMM & \cite{Calinon2016} & Adapt movements to new situations encountered by a robot automatically.\\
    \cline{4-4}
    \multirow{15}{*}{} & KMP & \cite{2019IJRR-KernelizedMovementPrimitives} & Enable robots to adapt learned skills to unforeseen obstacles and high-dimensional inputs through kernel-based trajectory learning and multi-coordinate representations.\\
    \cline{4-4}
    \multirow{15}{*}{} & {\color{black} BC /  --- --- } & \cite{2024RAL-identifyExpert} & Combine behavior classification and geometric augmentation to enable effective behavioral cloning from mixed-quality demonstration datasets.\\
    \cline{4-4}
    \multirow{15}{*}{} & {\color{black} BC / DIME} & \cite{2023icra-DIME} & {\color{black} Capture demonstrations from the real-time visual stream of a human operator's hand, utilize these demonstrations to teleoperate a robotic hand, and learn IL policies based on the robotic hand’s actions.}\\
    \cline{4-4}
    \multirow{15}{*}{} & {\color{black} BC / DIH-Tele} & \cite{2025TransactionsOnMechatronics-DIH-Tele} & {\color{black} Utilized proprioception, demonstration action sequence, and tactile information to train IL policies.
    }\\
    \cline{4-4}
    \multirow{15}{*}{} & {\color{black} BC / ViTacFormer} & \cite{heng-2025-vitacformer} & {\color{black} Utilized proprioception, demonstration action sequence, visual, and tactile information to train IL policies.}\\
    \cline{4-4}
    \multirow{15}{*}{} & {\color{black} BC / DexForce} & \cite{2025-RA-L-DexForce} & {\color{black} Utilize force-informed actions to train IL policies. These force-informed actions are computed based on the contact forces which measured during kinesthetic demonstrations.}\\
    \cline{4-4}
    \multirow{15}{*}{} & {\color{black} BC / DexSkills} & \cite{2024IROS_DexSkills} & {\color{black} Utilize tactile features acquired from the demonstrations to predict the on going primitives of the long-horizontal dexterous manipulation tasks.}\\
    \cline{4-4}
    \multirow{15}{*}{} & IRL & \cite{2021ICDL-InverseReinforcementLearning} & {\color{black} Learn the reward function using samples obtained from demonstrations of desired behaviors. Use statistical tools for random sample generation and reward normalization to reduce demonstration-dependent reward bias.}\\
    \cline{4-4}
    \multirow{15}{*}{} & GAIL & \cite{NIPS2016-AdvancesinNeuralInformation} & {\color{black} An adversarial imitation learning algorithm where a policy learns to mimic expert behavior by tricking a discriminator which tries to distinguish it from the expert.}\\
    \cline{4-4}
    \multirow{15}{*}{} & {\color{black} AIRL} & \cite{2021ICDL-InverseReinforcementLearning} & {\color{black} Recover policies and interpretable reward functions simultaneously within an adversarial learning framework. thereby providing strong transferability and generalization capability.}\\
    \cline{4-4}
    \cline{1-4}
    
    \multirow{17}{*}{\makecell{Reinforcement \\ Learning}} & Bi-DexHands & \cite{2024TPAMI-Bi-DexHands} & A novel simulator with two dexterous hands featuring 20 bimanual manipulation tasks and thousands of target objects, designed to match various levels of human motor skills based on cognitive science research.\\
    \cline{4-4}
    \multirow{17}{*}{} & Unidexgrasp++ & \cite{wan2023unidexgrasp++} & Object-agnostic, a universal policy for dexterous object grasping from realistic point cloud observations and proprioceptive information under a table-top setting. geometry feature of the task is utilized.\\
    \cline{4-4}
    \multirow{17}{*}{} & TOPDM & \cite{2021pmlr-SolvingChallengingDexterousManipulation} & A simple trajectory optimisation algorithm is presented and sub-optimal demonstrations generated through this optimisation algorithm are used for RL.\\
    \cline{4-4}
    \multirow{17}{*}{} & --- --- & \cite{2024ICRA-preTrain} & Predict the structural patterns in the environment through existing foundational vision networks without any fine-tuning.\\
    \cline{4-4}
    \multirow{17}{*}{} & --- --- & \cite{2024ICRA-Tactile-basedReinforcementLearning} & Utilize visual tactile information to train RL policies. Policies that obtained by RL can be generalized to unseen objects.\\
    \cline{4-4}
    \multirow{17}{*}{} & Tactile-AIRL & \cite{2024IROS-TactileActiveInference} & Active inference which integrates model-based techniques and intrinsic curiosity are introduced into the RL process. Model-based approach is utilized to imagine and plan appropriate actions through free energy minimization.\\
    \cline{4-4}
    \multirow{17}{*}{} & T-TD3 & \cite{2024TASE-T-TD3} & Involve tactile information, aim at deformable objects manipulation.\\
    \cline{4-4}
    \multirow{17}{*}{} & {\color{black} --- ---} & \cite{2025-RAS-cylindrical} & {\color{black} Leverage high-dimensional tactile information from interactions to train a DRL policy to control the pose of a thin cylindrical object in a tracking task.}\\
    \cline{4-4}
    \multirow{17}{*}{} & {\color{black} DexTouch} & \cite{2024-RAL-DexTouch} & {\color{black} Rely solely on tactile sensing to perform manipulation tasks, with vision excluded.}\\
    \cline{4-4}
    \multirow{17}{*}{} & {\color{black} VTAO-BiManip} & \cite{sun-2025-vtaobimanip} & {\color{black} leverage visual–tactile–action pretraining with object understanding to facilitate reinforcement learning and achieve human-like bi-manual manipulation.}\\
    \cline{4-4}
    \multirow{17}{*}{} & DartBot & \cite{2025TASE-DartBot} & Aim at achieve robust throwing skills for nonrigid relatively small objects.\\
    \cline{4-4}
    \multirow{17}{*}{} & {\color{black} DAPG} & \cite{Rajeswaran-RSS-18} & DRL is augmented with a small number of human demonstrations.\\
    \cline{4-4}
    \multirow{17}{*}{} & DexH2R & \cite{zhao2024dexh2rtaskorienteddexterousmanipulation} & Combine human hand motion re-targeting with a task-oriented residual action policy, improve task performance by bridging the embodiment gap between human and robotic dexterous hands.\\
    \cline{4-4}
    \multirow{17}{*}{} & --- --- & \cite{2025SR-DevelopmentOfCompositionality} & A brain-inspired neural network integrates vision, proprioception, and language through predictive coding to enhance robotic generalization of unlearned verb-noun compositions via sensorimotor learning.\\
    \cline{4-4}
    \multirow{17}{*}{} & {\color{black} LEGION} & \cite{Meng2025} & A Bayesian-inspired lifelong RL framework integrates language embeddings to enable robots to continuously accumulate and reuse knowledge from sequential tasks, advancing generalizable robotic intelligence.\\
    \hline
    
    \end{tabular}
} 
\end{threeparttable} 
\vspace{-5pt}
\label{IL and RL}
\end{table*}

\subsection{Imitation Learning}

IL can be divided into two subcategories. The first involves probabilistic modeling of human demonstration data using Gaussian Mixture Models (GMM) and Gaussian Mixture Regression (GMR). During skill reproduction, the trajectory is parameterized and optimized to match the learned probabilistic model, ensuring the reproduced motion retains human-like characteristics. Compared to deep learning and RL, this approach requires minimal training data and no prior knowledge while avoiding reliance on large neural networks, offering better mathematical explainability. Commonly used methods include Dynamic Movement Primitives (DMP)\cite{2013-neuralcomp-learningAttractor}, Stable Estimator of Dynamical Systems (SEDS)\cite{2011TRO-LearningStable}, Probabilistic Movement Primitives (ProMP)\cite{2013NIPS-ProbabilisticMovementPrimitives}, Task-Parameterized GMM (TPGMM)\cite{Calinon2016}, and Kernelized Movement Primitives (KMP)\cite{2019IJRR-KernelizedMovementPrimitives}. However, these methods are primarily suitable for simple trajectory reproduction and struggle with the interactive tasks that involve complex visual-tactile information. Their application has also been largely limited to single-task learning with robotic arms and two-fingered grippers, with few extensions to multi-fingered dexterous hands.

The second subcategory of IL employs deep learning to train policy networks that mimic human expert decisions. Unlike RL, this approach does not rely on environmental interaction or rewards but learns from pre-collected datasets. 
Common methods include Behavior Cloning (BC)\cite{2024RAL-identifyExpert, 2023icra-DIME, 2025TransactionsOnMechatronics-DIH-Tele, heng-2025-vitacformer, 2025-RA-L-DexForce, 2024IROS_DexSkills}, Inverse reinforcement learning (IRL) \cite{2021pmlr-Model-Based, 2021ICDL-InverseReinforcementLearning}, {\color{black} Adversarial Inverse Reinforcement Learning (AIRL) \cite{2021ICDL-InverseReinforcementLearning},} and Generative Adversarial Imitation Learning (GAIL) \cite{2021ICDL-InverseReinforcementLearning, NIPS2016-AdvancesinNeuralInformation}. {\color{black} Among these three methods, BC has generally demonstrated superior performance \cite{2024RAL-identifyExpert} and has therefore received considerable attention in recent years. In \cite{2023icra-DIME}, demonstrations for a nearest neighbor-based IL is generated by a teleoperation system. In this teleoperation system, a vision system is used to extract the movements of the operator's fingertips. These movements are then used to control the movements of a robotic hand. The method proposed in \cite{2023icra-DIME} demonstrates excellent performance in complex in-hand manipulation tasks. However, for these contact-rich tasks, no tactile information is incorporated. Incorporating tactile feedback could potentially further improve the learning outcomes. In \cite{2025TransactionsOnMechatronics-DIH-Tele}, tactile images along with the positional information are used as input to generate action sequences. In \cite{heng-2025-vitacformer}. except for the positional information and the tactile information, visual information are also used. The multi-modal representation enables IL for multi-fingered dexterous hands, enabling precise and adaptive manipulation. In \cite{2025-RA-L-DexForce}, the IL policies are trained on force-informed actions. These force-informed actions are computed based on the contact forces which measured during kinesthetic demonstrations. In \cite{2024IROS_DexSkills}, tactile features acquired from the demonstrations are used to predict the on going primitives of the long-horizontal dexterous manipulation tasks.}
A key limitation is that these methods only learn static behaviors from the demonstration data and cannot surpass human performance as RL might. Additionally, their performance degrades significantly when encountering states outside the training distribution, highlighting robustness issues.

\begin{figure*}[b]
    \centering
    \includegraphics[width=0.95\linewidth]{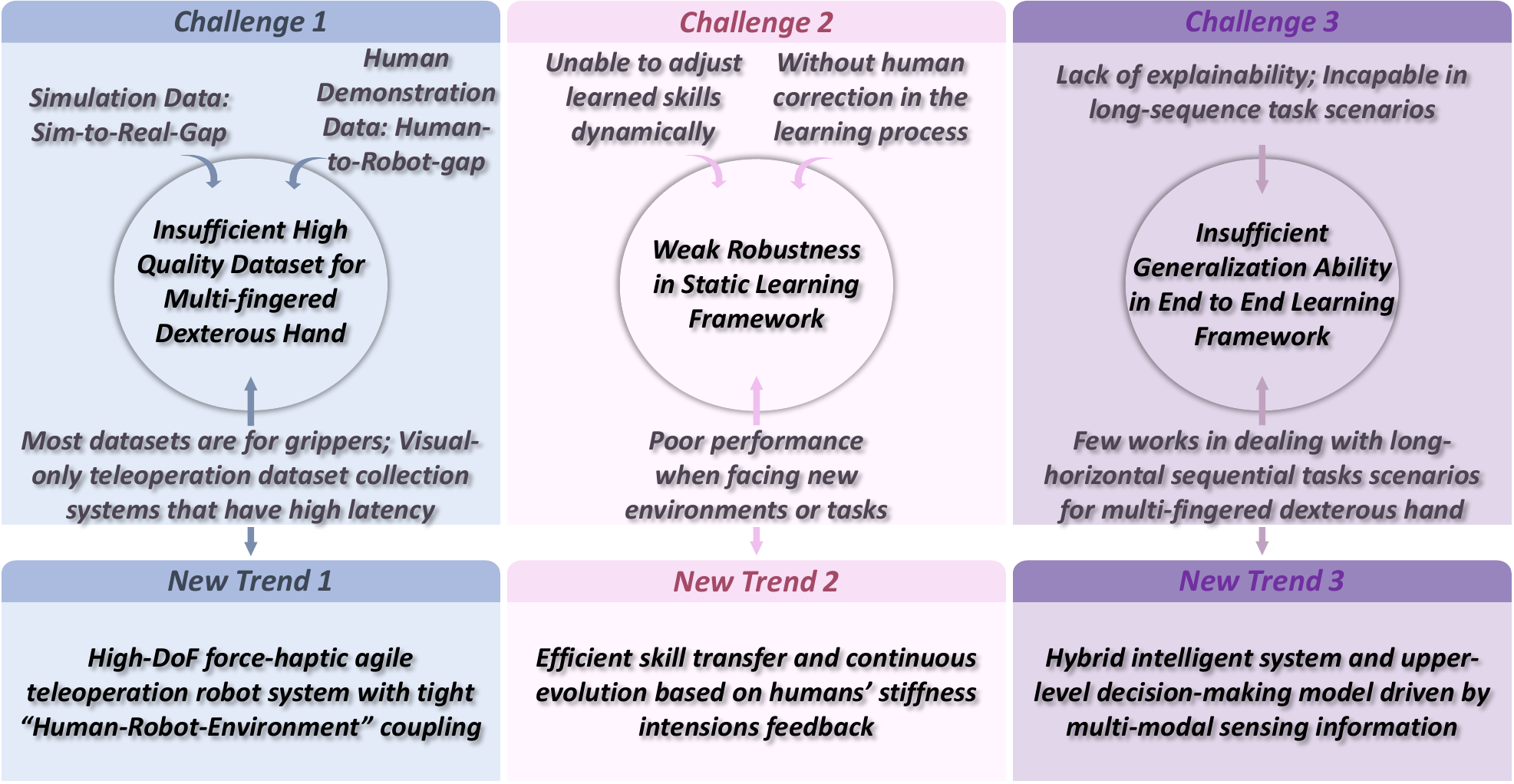}
    \caption{Challenges and new trends in robotic dexterous manipulation learning}
    \label{Fig_challenge_and_trend}
\end{figure*} 

\subsection{Reinforcement Learning}

RL has become the dominant approach for robotic dexterous manipulation skill learning, enabling agents to interact with their environment and refine policies through reward feedback. Recent researches have developed specialized simulation environments (e.g., Bi-DexHands for bi-manual manipulation)\cite{2024TPAMI-Bi-DexHands}, and RL methods for robotic control. Compared with traditional feedback control methods, RL has successfully tackled challenging manipulation skills\cite{wan2023unidexgrasp++, 2021IJRR-HowToTrain, 2021pmlr-SolvingChallengingDexterousManipulation}. However, applying RL to multi-fingered dexterous hands presents unique challenges due to the high degrees of freedom and complex contact dynamics\cite{2022FrontiersInNeurorobotics-Review}. Pure RL that only relies on agent-environment often suffers from sparse rewards and low sample efficiency, typically requiring millions of iterations to learn useful behaviors. The exploration process can also lead to incoherent or unsafe actions, making it difficult to achieve desired learning outcomes. To enhance learning efficiency, recent studies have utilized features extracted by pre-trained models to guide downstream robot tasks\cite{Radford2018ImprovingLU, 2024ICRA-preTrain}. The use of existing pre-trained models, combined with vast internet data, enables rapid and efficient learning of new tasks. In robotic RL, this pre-trained paradigm offers a key advantage: robots can acquire new skills with minimal human intervention by leveraging publicly available data and pre-trained models. 

Same as IL, tactile information can be quite useful. Su {\it et al.} employed RL to train a robot to pivot an object to a target orientation using only proprioceptive and tactile information\cite{2024ICRA-Tactile-basedReinforcementLearning}. Liu {\it et al.} proposed a novel framework called TactileAIRL for skill learning in robotic dexterous manipulation. Model-based techniques and intrinsic curiosity are integrated into the RL process, the method leverages vision-based tactile sensing to extract meaningful contact features. This design makes TactileAIRL scalable to many manipulation tasks learning involving tactile feedback\cite{2024IROS-TactileActiveInference}. Zhou {\it et al.} presented an RL framework called T-TD3 for stable grasping of deformable objects using tactile prior information\cite{2024TASE-T-TD3}. Aslam {\it et al.} presented a robot called DartBot that integrates tactile exploration and RL to achieve robust throwing skills in object transfer through throwing task\cite{2025TASE-DartBot}. {\color{black} Hu {\it et al.} \cite{2025-RAS-cylindrical} proposed leveraging high-dimensional tactile information from interactions between a thin stick and the fingertips of a three-fingered robotic hand to train a DRL policy for a tracking task. The policies were trained in simulation and successfully transferred to real-world experiments. Lee {\it et al.} \cite{2024-RAL-DexTouch} proposed a high-DoF dexterous robotic system with tactile-based blind manipulation capabilities. The system relies solely on tactile sensing to perform manipulation tasks, with vision excluded. Sun {\it et al.} \cite{sun-2025-vtaobimanip} proposed leveraging visual–tactile–action pretraining with object understanding to facilitate reinforcement learning and achieve human-like bi-manual manipulation.}

Apart from methods mentioned above, to promote learning efficiency of RL, recent studies have focused on the incorporating human prior knowledge into RL. Some studies leverage high-quality expert demonstrations to accelerate training, as seen in methods like DAPG mentioned by Rajeswaran {\it et al.} \cite{Rajeswaran-RSS-18}. Other studies, such as DexH2R\cite{zhao2024dexh2rtaskorienteddexterousmanipulation}, employ human-in-the-loop frameworks where operators provide corrective feedback during training. Similarly, UC Berkeley's interactive system also adopts this approach, during learning, the system continuously checks for potential corrections from human operators and updates the policy offline if adjustments are detected\cite{luo2025precisedexterousroboticmanipulation}. Research demonstrates that such human interventions enable robots to learn more effectively from mistakes, significantly improving their performance. While these approaches improve learning efficiency, current implementations rely on weakly coupled, discrete human supervision rather than continuous, tightly integrated guidance. 

Most RL methods focus on single-task learning, struggling with long-horizontal sequential tasks that require compositional skills. Long-horizontal sequential tasks consist of interconnected single-tasks. However, while RL focuses on final outcomes (e.g., reward maximization or goal achievement), it struggles to learn the transition conditions between them. Recent work has begun exploring the combinations of skills in long-horizontal sequential tasks, such as language-guided skill composition Meng {\it et al.} \cite{Meng2025} proposed the LEGION framework for lifelong learning, which leverages Bayesian nonparametric models and language embedding methods. This method enables robots to incrementally accumulate knowledge during continuous task execution. By effectively integrating and reusing acquired knowledge, the framework facilitates the solution of complex, long-term tasks. However, these advances remain limited to simple grippers, while the compositionality for skills on multi-fingered dexterous hand still largely unexplored. For multi-fingered dexterous hands, the Helix framework proposed to combine System 1 (Fast Reactive Control) and System 2 (Slower Deliberative Reasoning) to mimic the humans' cognitive systems. But how to implement this concept in practice remains challenging. How to build skill repository for multi-fingered dexterous hands, how to realize the combination of multiple skills remain open and require to be further studied.

\section{ Open Challenges and New Trends }
\label{Section-VI}

Endowing robots with human-like dexterous manipulation abilities has always been the substantial goal for researchers in the field of robotics. Since the early days of Unimate robots performing pre-programmed pick-and-place tasks in structured factory environments, robotic dexterous manipulation has evolved through three key stages: the mechanical programming stage, the closed-loop control stage, and the embodied dexterous manipulation stage. Despite significant progress in each stage, robotic dexterous manipulation capabilities are still far behind the ones of human beings, particularly for multi-fingered hands. Several key challenges, which are summarized and concluded in Fig. \ref{Fig_challenge_and_trend}, hinder the progress.

First, high-quality datasets for multi-fingered dexterous hand is insufficient. 
1) Datasets constructed solely through "robot-environment" interactions exhibit low information entropy, often requiring millions of iterations to learn useful knowledge. For instance, datasets generated in the simulation environments inherently face the Sim-to-Real gap, limiting their real-world applicability. 2) Except for dataset generated in simulation environments, alternative approaches such as collecting data through human demonstrations, introduce the Human-to-Robot gap. Despite their richness, these datasets have to struggle with kinematic and dynamic mismatches between human and robotic systems, which makes the skill transfer from human to robots be difficult. 3) Most existing manipulation skill datasets are collected based on simple two-fingered grippers. While the datasets collected with high-DoF multi-fingered dexterous hands are insufficient. 4) Existing teleoperation-based data collection systems can address the Sim-to-Real and Human-to-Robot gaps. However, they also exhibit several limitations: i) Existing systems rely solely on visual feedback, resulting in weak "Human-Robot" coupling. This approach fails to incorporate critical force and tactile feedback, which severely limits the incorporations of humans' haptic experience to the manipulation tasks. ii) The current latencies for existing systems is generally at several milliseconds, which are slow for many delicate manipulations that required agilities. 

{\color{black} Considering the above challenges, achieving precise capture of multi-modal datasets for multi-fingered dexterous robotic hands remains a non-trivial task. A promising trend for data collection is to develop a high-DoF haptic agile teleoperation robot system with tight ``Human-Robot-Environment" coupling. Some studies have already begun to explore this potential solution. 
In \cite{2020-Advanced-Robotics-fingertip-haptic-feedback}, a teleoperation system with haptic feedback is developed. In this system, a custom-designed 10-DoFs robotic hand with tactile sensing is teleoperated to perform tasks. During task execution, pressure sensors on the fingertips of the robotic hand collect contact force data, while haptic feedback actuators located on the operator’s fingertips display the contact forces. However, the robotic hand used has only 10 DoFs, fewer than a truly dexterous hand (which typically has more than 20 DoFs, e.g., the Shadow Hand). Moreover, in \cite{2024IROS_DexSkills}, a high-DoF teleoperation system is built to gather training data for robot skill learning. During data collection, 
tactile information, joint angle states, and end-effector states are recorded. 
Such a system can address both the Sim-to-Real and Human-to-Robot gaps in multi-fingered dexterous manipulation. In addition, except for visual modal, integrating other modal sensing (e.g., force, tactile) can enrich contact information, thereby improving manipulation performance. Therefore, we believe the haptic agile teleoperation systems are highly promising for building high-quality datasets for multi-fingered dexterous hands. }

Second, the robustness of the static learning framework is weak. 1) 
{\color{black} Current manipulation skill learning frameworks that based on RL/IL can only statically acquire or extract manipulation skills through trial-and-error exploration or from existing datasets,
unable to dynamically adjust the learned skills.} 2) Integrating humans effectively into the learning loop also remains a challenge. In existing methods, human supervision and guidance are weakly coupled, which makes it difficult to accurately reflect humans' satisfaction with the skill learning. 3) Owning to the mechanisms of RL/IL, when the environments or tasks are beyond the dataset's coverage, their performance would degrade significantly. {\color{black}Thus to overcome the limitations of static learning frameworks, a promising trend is to incorporate humans  tightly into the learning loop, enabling robots to acquire more robust manipulation skills. As mentioned above, using human's guidance to promote learning has already been studied and emerged as a trend.
In \cite{2023-IROS-Primitive-Skill-Based}, human's guidance is indicated with discontinuous binary signals. These binary signals are generated via keyboard key pressing. For instance, if the operator satisfies with the current performance, he or she will give a positive signal. On the contrary, if the performance is not desired, a negative signal will be presented. In \cite{wang-2021-skill-preferences-learning-extract}, a human is asked to provide preferences between different skill sequences. Each preference is also indicated by a binary signal, where the preferred sequence is labeled as positive and the other as negative. These preferences are then used to train a reward model, enabling it to assign higher reward values to the human-preferred sequence. The trained reward model substitutes the hand-crafted reward function in a RL framework. Consequently, the RL framework learns to automatically select the skill sequence that a human is most likely to prefer. 

Although using discontinuous binary signals as human's feedback has achieved great successes. They are weakly coupled and have the natural disadvantage of not being able to express the extent of human satisfactoriness. On the other hand, due to the convenience of acquisition and rich representation of human muscle activity. The stiffness of human arm endpoint has already used in many applications \cite{2012-IJRR-Tele-impedance, 2018-IJRR-Reduced-complexity}. However, using stiffness to indicate human's guidance and preference in the learning process is still understudied. Thus we believe that using stiffness to represent human's guidance and preference to enhance skills learning and facilitate skills transfer from human to robots is a worthwhile research direction and a promising trend in the learning processes.} 

Third, the generalization ability of end-to-end learning framework is insufficient. Many robot manipulation skill learning predominantly adopts an end-to-end learning framework. However, this learning framework suffers from issues such as high task-coupling and lack of explainability. These issues make the end-to-end learning framework struggle to adapt to long-horizontal sequential tasks scenarios. 

Scientific researches \cite{2007-Human-Movement-Science-Hierarchical, 2019-NC-Hierarchical} show that humans do not follow an end-to-end framework in learning to perform long-horizontal sequential tasks. Instead, when facing long-horizontal sequential tasks, humans naturally perform the task decomposition and then execute the primitive skills automatically and unconsciously. Inspired by this mechanism, a novel robotic manipulation learning framework, Helix framework, is proposed. This framework combines System 1 (Fast Reactive Control) and System 2 (Slower Deliberative Reasoning).
{\color{black} Previous studies have already explored hierarchical frameworks which are similar to the Helix framework to address long-horizon tasks. Some studies used two-level hierarchical frameworks. For instance, in \cite{sun-2024-hierarchical-hybrid-learning-long-horizon}, the proposed framework integrates a low-level primitive library of continuously parameterized skills with a high-level policy. The low-level library contains essential skills for assembly tasks, such as grasping and inserting, implemented using both RL and model-based controllers. The high-level policy, learned via imitation learning from a small set of demonstrations, selects appropriate primitive skills and instantiates them with continuous input parameters. On the other hand, some studies used three-level hierarchical frameworks \cite{2025-TASE-Learn-Gen-Plan, li-2025-atomic-skill-library-construction}. It is noteworthy that different studies may adopt different definitions of the functionalities of these levels. In \cite{2025-TASE-Learn-Gen-Plan}, the lowest level conducts primitive learning, the middle level uses the learned primitives to generate robotic skills, and the highest level adopts a Vision-Language Model (VLM) to decompose long-horizontal tasks into separate robotic skills, meanwhile, these robotic skills are planed and formed in to an action sequence for execution. In \cite{li-2025-atomic-skill-library-construction}, at the highest level, a Vision-Language-Planning (VLP) agent performs task decomposition and planning. This agent integrates semantic instructions, visual observations, and spatial intelligence to generate a sequential plan of sub-tasks. Subsequently, at the middle level, a semantic abstraction module maps each sub-task into a generalized atomic skill definition. Finally, at the lowest level, a Vision-Language-Action (VLA) model is fine-tuned to execute these atomic skills, thereby completing the original task. These hierarchical frameworks have demonstrate well performance in the long-horizontal tasks. 

However, above studies encountered a common limitation that the end-effectors are mainly two-fingered grippers. Neither dexterous high-DoF robotic hands nor haptic information are incorporated.} Research on compositional skills for multi-fingered dexterous hands is still in its early days. 
{\color{black} 
Thus we believe that integrating the task reasoning and fast control capabilities of hierarchical frameworks with high-DoF haptic hands is promising. Such integration enables dexterous object manipulation and could significantly improve overall system performance. Therefore, we consider the incorporation of multi-fingered dexterous hands and multi-modal sensing into hierarchical learning frameworks to be a highly promising trend for future research.}

\section{Conclusion}
\label{Section-VII}
This survey has systematically reviewed the state-of-the-art advancements and challenges in dexterous manipulation. Historical development of robotic manipulation, key bottlenecks in current research, contemporary data collection methods and learning frameworks, along with their respective advantages, and limitations are discussed sequentially. Based on the state-of-the-art advancements and challenges, we believe that this survey captured the most important features in robotic manipulation. At the same time, some future thoughts are made: Endowing multi-fingered dexterous hands with human-like manipulation capabilities can enhance the potential to perform complex tasks for robots, especially humanoid robots. This advancement would simultaneously promote robotic and AI technologies and then significantly enhance productivity. Meanwhile, expanding researches on skill learning methods for multi-fingered dexterous hands could lead to a deeper understanding of human brain decision-making functions. Such progresses would greatly accelerate robotic development and bring us closer to achieving real AI technologies and embodied intelligence.






\bibliography{ref}

 




\vfill

\end{document}